\documentclass[10pt,twocolumn,letterpaper]{article}

\usepackage{cvpr}              

\usepackage{graphicx}
\usepackage{amsmath}
\usepackage{amssymb}
\usepackage{booktabs}
\usepackage[table,xcdraw]{xcolor}
\usepackage{graphicx}

\usepackage[accsupp]{axessibility}
\usepackage[pagebackref,breaklinks,colorlinks]{hyperref}

\usepackage[capitalize]{cleveref}
\crefname{section}{Sec.}{Secs.}
\Crefname{section}{Section}{Sections}
\Crefname{table}{Table}{Tables}
\crefname{table}{Tab.}{Tabs.}

\def\cvprPaperID{2511} 
\def\confName{CVPR}
\def\confYear{2022}

\begin{document}

\title{Cross Domain Object Detection by Target-Perceived Dual Branch Distillation}

\author{Mengzhe He$^{1,3}$,~~ Yali Wang$^{\ast1,6}$,~~ Jiaxi Wu$^{5}$,~~ Yiru Wang$^{2}$,\and  Hanqing Li$^{2}$,~~ Bo Li$^{2}$,~~ Weihao Gan$^{2,4}$,~~ Wei Wu$^{2,4}$,~~ Yu Qiao$^{\dagger1,4}$ \\
{\footnotesize\centering$^{1}$  ShenZhen Key Lab of Computer Vision and Pattern Recognition, Shenzhen Institute of Advanced Technology, Chinese Academy of Sciences}~~~~~\\
{\small\centering$^{2}$ SenseTime Research}~~~~~
{\small\centering$^{3}$ University of Chinese Academy of Science}~~~~~
{\small\centering$^{4}$ Shanghai AI Laboratory, Shanghai, China}
\\
{\small\centering$^{5}$ Beihang University}~~~~~
{\small\centering$^{6}$ SIAT Branch, Shenzhen Institute of Artificial Intelligence and Robotics for Society}\\
\tt\small \{hemz,yl.wang,yu.qiao\}@siat.ac.cn, wujiaxi@buaa.edu.cn\\ {\tt \small\{lihanqing,libo,wuwei\}@senseauto.com, \{wangyiru,ganweihao\}@sensetime.com}}

\maketitle
\footnote{${^*}$ Equal contribution.\ \  ${\dagger}$ Corresponding author.}
\begin{abstract}
Cross domain object detection is a realistic and challenging task in the wild. It suffers from performance degradation due to large shift of data distributions and lack of instance-level annotations in the target domain. Existing approaches mainly focus on either of these two difficulties, even though they are closely coupled in cross domain object detection. To solve this problem, we propose a novel Target-perceived Dual-branch Distillation (TDD) framework.  By integrating detection branches of both source and target domains in a unified teacher-student learning scheme, it can reduce domain shift and generate reliable supervision effectively. In particular, we first introduce a distinct Target Proposal Perceiver between two domains. It can adaptively enhance source detector to perceive objects in a target image, by leveraging target proposal contexts from iterative cross-attention. Afterwards, we design a concise Dual Branch Self Distillation strategy for model  training, which can progressively integrate complementary object knowledge from different domains via self-distillation in two branches. Finally, we conduct extensive experiments on a number of widely-used scenarios in cross domain object detection. The results show that our TDD significantly outperforms the state-of-the-art methods on all the benchmarks. Our code and model will be available at \href{https://github.com/Feobi1999/TDD}{here}.
\vspace{-0.3cm}
\end{abstract}

\section{Introduction}
\label{sec:intro}
Object detection has achieved remarkable success with the help of advanced deep neural networks \cite{cai2018cascade,7410526,8237584,he2015spatial,lin2017focal,ren2015faster,7780460,redmon2017yolo9000,redmon2018yolov3,shen2017dsod}. However, it still faces challenges in realistic applications such as autonomous driving and mobile robots, where data variance is often large due to various conditions of weather, illumination, object appearance, etc. Hence, cross-domain object detection has attracted lots of attention in recent years. In general, there are two difficulties in this problem. First, object detection is more vulnerable to domain shift. The main reason is that, object detection focuses on instance-level prediction, which is more sensitive to object variance in various image styles and contents. Second, object annotations are more expensive and labor-intensive to get, causing the scarcity of discriminative object supervision in a new domain. Both of them inevitably deteriorate the detection performance in target domain. 
\begin{figure}[t]
  \centering
  \includegraphics[width=1\linewidth]{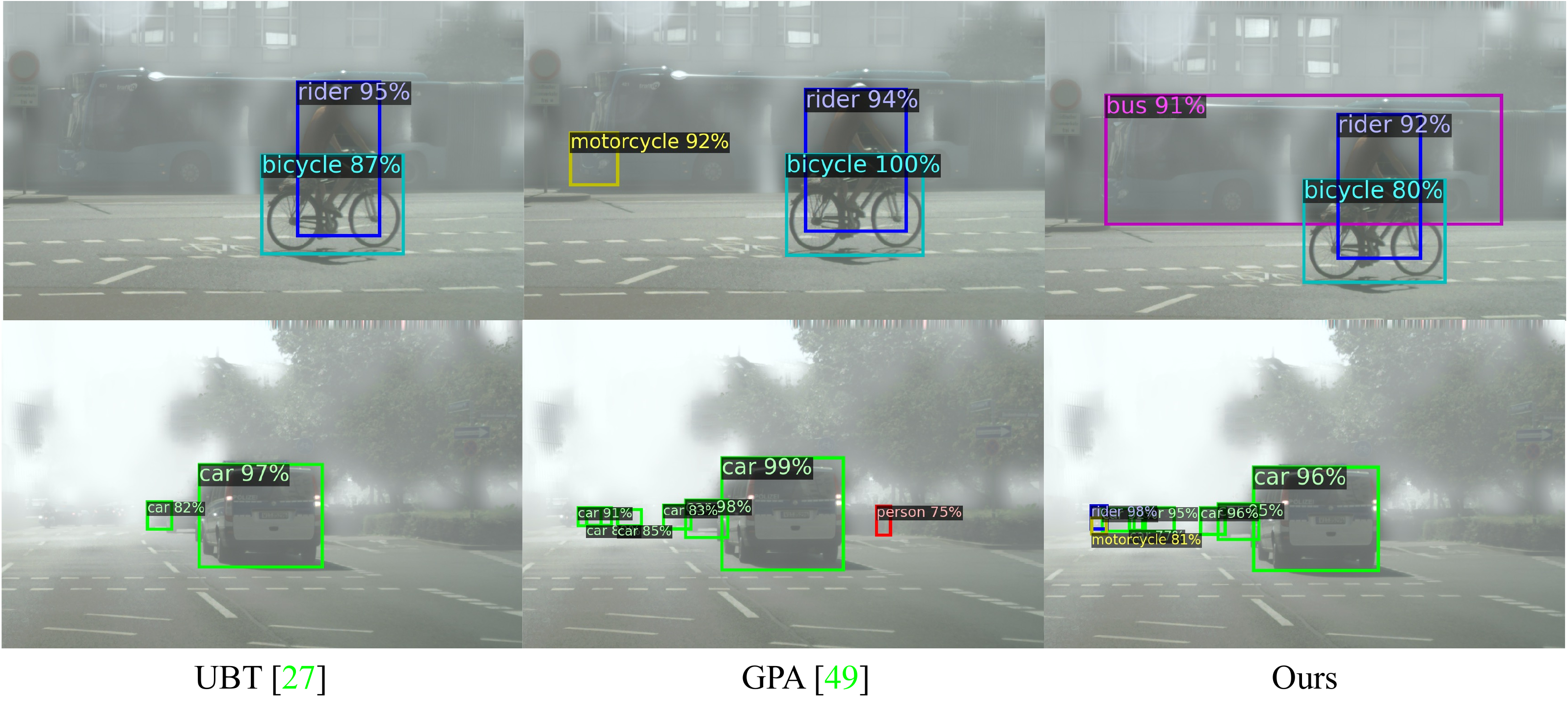}
   \vspace{-0.6cm}
   \caption{Two typical examples of detection results on the adverse weather conditions adaptation experiments with different methods. Semi-supervised method UBT\cite{liu2021unbiased} lacks awareness of objects in the fog.  Adversarial based GPA\cite{9157427} attempts to exploit the objects in the fog but gives some wrong predictions, such as the motorcycle in the first row and the person in the second row. Our methods can predict the boxes and categories more accurately.}
   \label{fig:intro}
   \vspace{-0.6cm} 
\end{figure}

Recently, several approaches have been proposed for cross-domain object detection\cite{8578450,8954025,Rezaeianaran_2021_ICCV,8954336,9157427}. Unfortunately, most of them focus on either domain shift or label deficiency, which limits their power in cross domain object detection. For example, domain adaption approaches\cite{8578450,8954336,9157427} propose to reduce domain shift via adversarial training. Besides of unstable model optimization, the discrimination ability of the network is limited in such adversarial design. As shown in Figure \ref{fig:intro}, adversarial based GPA\cite{9157427} tends to produce wrong predictions on the regions where the target domain characteristics are significant. Alternatively, self-training based approaches\cite{8953637,10.1007/978-3-030-58586-0_19,9008383,9010241,10.1007/978-3-030-58523-5_6} study the problem from the viewpoint of semi-supervised learning, and propose to generate pseudo object supervision via label distillation. In this way, many advanced semi-supervised methods can be transferred to this task. However, these approaches are often insufficient to deal with the complex domain shifts. In Figure \ref{fig:intro}, it is difficult for a semi-supervised method like UBT\cite{liu2021unbiased} to aware objects in the target domain. Hence, both types of solutions are unsatisfactory in cross domain object detection.

Based on these discussions, we propose a novel Target-perceived Dual-branch Distillation (TDD) framework, which can effectively tackle domain shift and label deficiency via object perception and knowledge distillation in a concise dual-branch detection network. Specifically, our network consists of a source-adaptive branch and a target-like branch, both of which are elaborately designed to be target-oriented for domain shift reduction. For the source-adaptive branch, we introduce a distinct Target Proposal Perceiver, which leverages iterative cross-attention to discover target-domain contexts for each proposal. As a result, it can adaptively enhance source branch to perceive objects in the target domain image. For the target-like branch, we transfer source images into target-like images. Via training this branch with these labeled images, we can learn discriminative object knowledge of target domain reliably. Finally, we design a concise Dual Branch Self Distillation strategy for network training. It is a tailored mean-teacher style framework to generate pseudo annotations of target images from both source-adaptive and target-like branches. Through three well-designed training steps, namely joint-domain pretraining,  cross-domain distillation and dual-teacher refinement, we can progressively integrate complementary object knowledge from different domains to boost cross domain object detection. 

In summary, this paper has the following contributions. First, we develop a novel Target-perceived Dual-branch Distillation (TDD) framework, which leverages two distinct detection branches to address both domain shift and label deficiency in a unified teacher-student learning manner. Second, we introduce a smart Target Proposal Perceiver module, which can adaptively guide source detection branch to perceive target domain objects, via cross-attention-style transformer on proposal contexts. Finally, we conduct extensive experiments on a number of widely-used benchmarks and our TDD outperforms the state-of-the-art methods with a large margin.
\section{Related Work}
\textbf{Object detection.} Object detection is one of the fundamental  tasks in computer vision. Boosted by the strong representation ability of deep neural network, object detection has obtained  a promising  performance in recent  years.  Previous work can be roughly categorized into two-stage\cite{cai2018cascade,7410526,8237584,he2015spatial,ren2015faster} and one-stage\cite{7780460,redmon2017yolo9000,redmon2018yolov3,shen2017dsod} detectors. Recently, some anchor-free\cite{duan2019centernet,tian2019fcos,yang2019reppoints,zhang2020bridging} and transformer\cite{carion2020end,wang2021pyramid,zhu2020deformable} based methods also stand out in the detection task. 

\textbf{Cross domain object detection.} \cite{8578450}first propose image and instance level domain classifiers to implement feature alignment in an adversarial manner. Following this, \cite{8954336}impose a strong-weak alignment strategy to the local and global features  respectively.\cite{9010003} and \cite{9022084} employ multi level domain feature alignment. \cite{9157382}exploit the categorical consistency between image-level and instance-level prediction with the help of a multi-label classification model. \cite{10.1007/978-3-030-58545-7_42} propose a center-aware feature alignment method to allow the discriminator to focus on features coming from the object region. Some other works\cite{10.1007/978-3-030-58586-0_19,8954025,Rezaeianaran_2021_ICCV,10.1007/978-3-030-58621-8_24,8953252} add additional constraint during the adversarial learning stage. \cite{Zhang_2021_CVPR,9156464} emphasis the different strategies to deal with foreground and background features. \\
\indent Another mainstream method\cite{8953637,10.1007/978-3-030-58586-0_19,9008383,9010241,10.1007/978-3-030-58523-5_6} is dedicated to solving the problem of inaccurate label in target domain.\cite{9008383}retrain the object detector using the original labeled data and the refined machine-generated annotations in the target domain. \cite{8953637} study the problem from the viewpoint of semi-supervised learning and integrate the object relations into the measure of consistency cost between teacher and student modules. \cite{Deng_2021_CVPR}propose a cross-domain distillation method which utilizes both the source-like and target-like images. It uses soft label and instance selection to heal the model bias in Mean-Teacher. Different from \cite{Deng_2021_CVPR}, our method proposes a dual-branch framework with a cross-domain perceiver for teacher-student mutual learning. \\ 
\indent \textbf{Semi-supervised object detection.} Semi-supervised object detection attempts to solve the problem when there are only a part of annotations for the train set. In this setting \cite{jeong2019consistency} propose a consistency-based method, enforcing the  predictions consistency between an input image and its flipped version. \cite{sohn2020simple} pre-train a detector using a small amount of labeled data and generate pseudo-labels on unlabeled data to fine-tune the pre-trained detector. \cite{liu2021unbiased} propose to use strong and weak augmentations to improve the mean-teacher method and can get more accurate pseudo labels by EMA training. Those methods can be easily applied to the cross domain object detection problem owing to the similar data setting. But they did not take the domain difference into consideration, which limited their detection performance unavoidably.

\begin{figure*}[t]
  \begin{center}
  \includegraphics[width=0.8\linewidth]{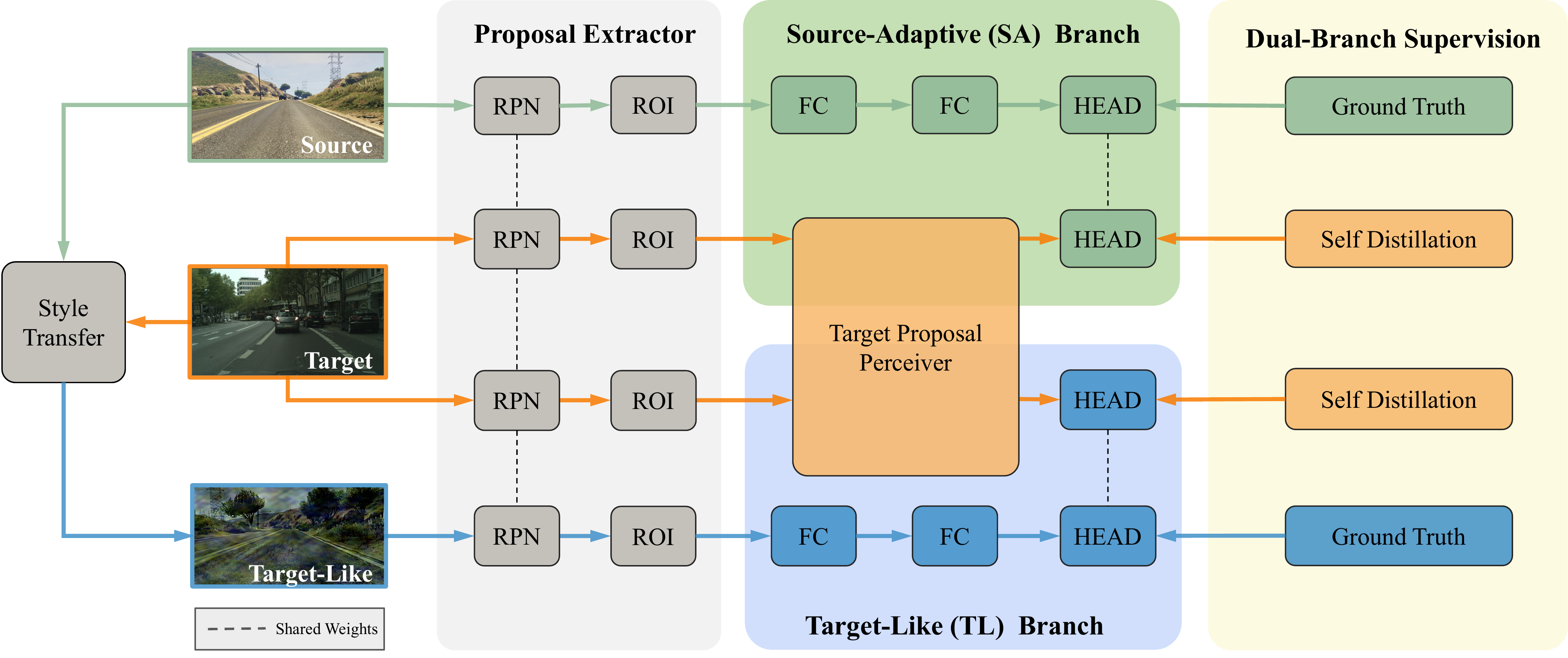}
 \end{center}
    \vspace{-0.4cm} 
   \caption{An overview of Target-perceived Dual-branch Distillation framework. To simplify the description, ROI refers to the operation to get proposal features of each image. First, a source domain image is transferred into target-like domain. All of the images from three domains are fed to a shared proposal extractor to get proposals and proposal features. Then, the proposal  features of source and target-like images are used to train corresponding branches with supervision of ground truth. Moreover, we feed the proposal features of a real target domain image into both branches, for learning object knowledge from both source and target-like domains. As the images from target domain are not annotated, the model is optimized by self-distillation.}   \label{framework}
   \vspace{-0.5cm}
\end{figure*}
\section{Proposed Methods}
\subsection{Overview}
As shown in Figure \ref{framework}, we propose a novel Target-perceived Dual-branch Distillation framework (TDD), which tackles domain shift and label deficiency together in cross domain object detection task.

First, we introduce a style transfer module from the aspect of input image. It is used to transfer source images into style that is close to target domain. In this case, we can bridge the domain gap by such target-like domain. Moreover, since target-like images inherit label annotations from the corresponding source images, they can be used as extra object supervision in the target-like domain. In this paper, we mainly use a concise and effective Fourier transform\cite{9157228} method as this module.

Second, we design a novel dual branch detection network from the aspect of model architecture. Via such design, we can effectively extract complementary object knowledge from different domains to boost object detection on the target images. Basically, our network consists of a shared proposal extractor and two individual detection branches. The former allows us to construct domain-invariant feature space of all the images for domain generalization, while the latter preserves domain-specific object characteristics of each image for domain discrimination. Specifically, two detection branches are Source-Adaptive (SA) and Target-Like (TL) branch respectively. We feed the proposals of source images to train the SA branch, while feeding the proposals of target-like images to train the TL branch. Moreover, the proposals of a real target image are sent into both branches, for learning object knowledge from both source and target-like domains. However, source domain may be significantly different from target domain. In this case, the proposals of a target image cannot be detected accurately in the SA branch, without any target-oriented guidance. To tackle this problem, we design a novel Target Proposal Perceiver. Inspired by perceiver in\cite{DBLP:journals/corr/abs-2103-03206}, it smartly uses iterative cross attention between proposal features in two branches. In this case, we leverage contextual proposals of TL branch as guidance, which can effectively guide SA branch to perceive object proposals in the target domain. We will explain the details of this module in \ref{sub3.2}.

Finally, we introduce a concise dual-branch self-distillation approach from the aspect of supervision. As introduced before, all the images do not have any annotations in target domain. Hence, it is critical to generate reliable supervision in this domain. Thanks to our dual-branch network, we can construct discriminative pseudo labels of each target image from the cooperative SA and TL branches. To effectively leverage these pseudo labels, our self-distillation is based on teacher-student mutual learning, which can dynamically adjust teacher in the training procedure to progressively boost target-domain supervision of our two branches. We will explain the details in \ref{sub3.3}.

\subsection{Target Proposal Perceiver}\label{sub3.2}
As discussed in our TDD framework, we feed proposal features of each target-domain image respectively into SA and TL branches, for learning object knowledge from both domains. 
However, SA branch is not good at exploiting  objects from these features due to the large shift between the source and real target domain. 
To guide SA branch to discover target domain objects, we propose a novel Target Proposal Perceiver between SA and TL branches, which can progressively exploit object contexts in the TL branch to enhance proposal features in the SA branch.

Note that, we inherit the name of Perceiver from \cite{DBLP:journals/corr/abs-2103-03206}, since our motivation is also to mimic humans and other animals to take in data from many sources and integrate it seamlessly. But different from the generic Perceiver\cite{DBLP:journals/corr/abs-2103-03206} architecture, our Target Proposal Perceiver is elaborately tailored for cross-domain object detection, by using Transformer-style cross attention iteratively to reduce domain shift in the instance level.

\begin{figure}[t]
  \centering
  \includegraphics[width=0.9\linewidth]{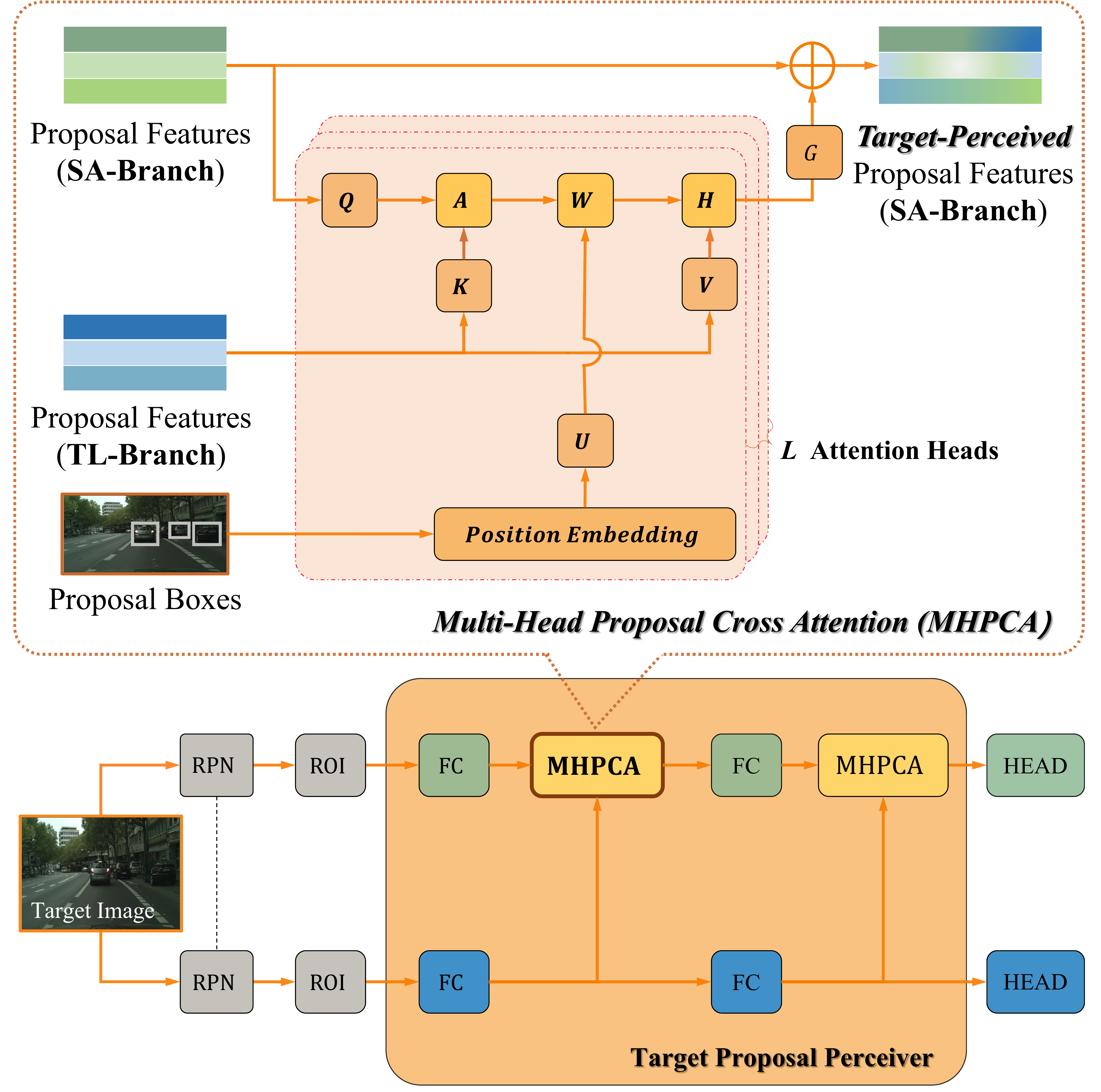}
   \caption{The structure of our Target Proposal Perceiver. The cross attention between SA and TL is explored to help source branch perceive target domain objects.}
   \label{transfomer}
   \setlength{\abovecaptionskip}{-0.5cm} 
    \vspace{-0.5cm} 
\end{figure}

As shown in Figure \ref{transfomer}, we feed a target-domain image $\mathbf{X}^{t}$ into proposal extractor, and generate its proposal features $\mathbf{P}^{t}$. Subsequently, we put these proposal features respectively into SA and TL branches, where Target Proposal Perceiver leverages cross attention to process them in the following,
\vspace{-0.2in}
\begin{align}
&\boldsymbol{\Phi}_{SA} = \mathcal{F}_{SA}(\mathbf{P}^{t}),\label{eq:perceiver1}\\
&\boldsymbol{\Phi}_{TL} = \mathcal{F}_{TL}(\mathbf{P}^{t}),\label{eq:perceiver2}\\
&\boldsymbol{\Psi}_{SA} =\text{MHPCA}(\boldsymbol{\Phi}_{SA}, \boldsymbol{\Phi}_{TL}).\label{eq:perceiver3}
\end{align}
\vspace{-0.2in}

\begin{figure*}[t]
  \begin{center}
  \includegraphics[width=0.75\linewidth]{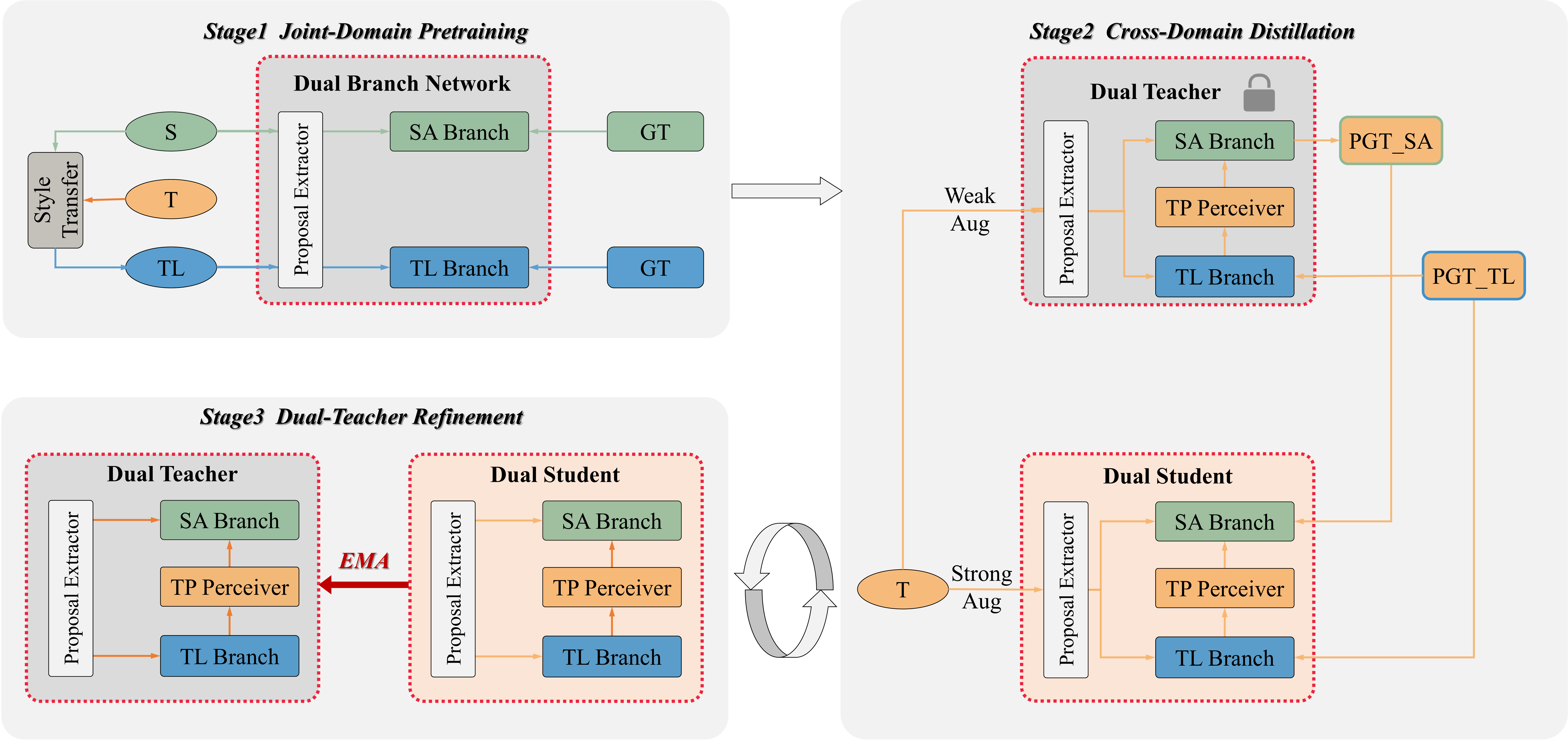}
   \caption{The whole training process of our Dual-Branch Self Distillation models. First, in the joint-domain pretraining stage, we pretrain our dual-branch network jointly, by multi-task learning on the labeled images of both source and target-like domains. Second, in the cross-domain distillation stage, we feed a target-domain image into the fixed and well-trained teacher,  which can generate pseudo object annotations from both SA and TL branches. Finally, to generate more stable pseudo annotations, we refine teacher gradually from student via Exponential Moving Average (EMA).
   }
   \label{trainingprocess}
  \end{center}
  \vspace{-0.8cm} 
\end{figure*}First, to extract object knowledge from both SA and TL branches,
we use the FC layer $\mathcal{F}_{SA}(.)$ and $\mathcal{F}_{TL}(.)$ to encode $\mathbf{P}^{t}$ as source features $\boldsymbol{\Phi}_{SA}$ and target-like features $\boldsymbol{\Phi}_{TL}$ in Eq. (\ref{eq:perceiver1})-(\ref{eq:perceiver2}).
Second, we introduce a novel Multi-Head Proposal Cross Attention (MHPCA) between $\boldsymbol{\Phi}_{SA}$ and $\boldsymbol{\Phi}_{TL}$ in Eq. (\ref{eq:perceiver3}).
This allows us to leverage target-like proposal features $\boldsymbol{\Phi}_{TL}$ as context guidance,
for enhancing source proposal features $\boldsymbol{\Phi}_{SA}$ to perceive objects in the target image.

\textbf{Proposal Cross Attention}.
Specifically, our MHPCA is a concise Transformer style with Query-Key-Value. In each cross attention head, we use FC layers to encode $\boldsymbol{\Phi}_{SA}$ as Query, and encode $\boldsymbol{\Phi}_{TL}$ as Key and Value. The similarity between Key and Query is used to discover affinity between $\boldsymbol{\Phi}_{SA}$ and $\boldsymbol{\Phi}_{TL}$. Then,
we use such affinity as guidance to aggregate target-like features $\mathcal{V}(\boldsymbol{\Phi}_{TL})$ as cross-domain contexts for the SA branch.
\vspace{-0.05in}
\begin{equation}
\mathbf{H}_{TL} = \mathcal{W}(\mathcal{Q}(\boldsymbol{\Phi}_{SA}),\mathcal{K}(\boldsymbol{\Phi}_{TL}))\cdot\mathcal{V}(\boldsymbol{\Phi}_{TL}),\label{eq:pca}
\end{equation}
where
Query, Key and Value are respectively $\mathcal{Q}(\boldsymbol{\Phi}_{SA})$, $\mathcal{K}(\boldsymbol{\Phi}_{TL})$ and $\mathcal{V}(\boldsymbol{\Phi}_{TL})$.
The affinity function is $\mathcal{W}$. Typically, scaled dot-product is used as $\mathcal{W}$ in transformer\cite{vaswani2017attention}, $
\mathbf{A}_{i,j}={\mathcal{Q}_{i}(\boldsymbol{\Phi}_{SA})\cdot\mathcal{K}_{j}^{\top}(\boldsymbol{\Phi}_{TL})}/{\sigma},
$
where $\sigma$ is a scale parameter that is root square of the dimension of a query feature vector.
However, we consider an object detection problem, where spatial position information can be important to describe similarity between proposals. In this work, the geometry weight in \cite{hu2018relation} is used to describe positional similarity between any two proposal boxes. We use this geometry weight $\mathbf{U}$ to enhance feature similarity $\mathbf{A}$ and describe proposal affinity in Eq. (\ref{eq:pca}) via a weighted formulation of softmax, 
i.e.,
$\mathcal{W}(\mathcal{Q}(\boldsymbol{\Phi}_{SA}),\mathcal{K}(\boldsymbol{\Phi}_{TL}))=\mathbf{W},$
\begin{equation}
\mathbf{W}_{i,j}
=\frac{\mathbf{U}_{i,j}\cdot\exp(\mathbf{A}_{i,j})}{\sum_{k=1}^{K}\mathbf{U}_{i,k}\cdot\exp(\mathbf{A}_{i,k})},\label{eq:affinityall2}
\end{equation}
where $\mathbf{W}_{i,j}$ refers to affinity score between proposal $i$ in SA branch and proposal $j$ in TL branch.

\textbf{Iterative MHPCA.} After obtaining target-like contexts $\mathbf{H}_{TL}$ from each cross attention head, we use FC layer $\mathcal{G}(.)$ to summarize all these contexts from $L$ attention heads to construct MHPCA, denoted as 
$
\boldsymbol{\Psi}_{SA}
=\boldsymbol{\Phi}_{SA}+\mathcal{G}([\mathbf{H}_{TL}^{(1)},...,\mathbf{H}_{TL}^{(L)}]).
$
In this case, we enhance source proposal features $\boldsymbol{\Phi}_{SA}$ into target-perceived ones $\boldsymbol{\Psi}_{SA}$,
which allows SA branch to be aware of related object contexts in the target image. Additionally, we perform MHPCA in an iterative manner, by which our Target Proposal Perceiver can progressively exploit target-like proposal contexts from TL branch to boost learning capacity of SA branch. Typically, there are two FC layers to encode proposal features in Faster RCNN. Hence, we iteratively use MHPCA twice in our design, as shown in Figure \ref{transfomer}.

\subsection{Dual-Branch Self Distillation}\label{sub3.3}

After introducing our network, we explain how to train it for cross domain object detection. As mentioned before, the images are unlabeled in the target domain. Hence, it is critical to generate reliable pseudo annotations of these images for effective training.
To achieve this goal,
we design a generic Dual-Branch Self Distillation approach,
which can generate pseudo labels from both SA and TL branches to cooperatively boost our detection network via self-training.
Specifically,
it is based on the general procedure of teacher-student mutual learning\cite{tarvainen2017mean,liu2021unbiased},
but with elaborate designs for cross domain object detection.
As shown in Figure \ref{trainingprocess},
it consists of three key stages, i.e., Joint-Domain Pretraining, Cross-Domain Distillation, and Dual-Teacher Refinement.

\textbf{Joint-Domain Pretraining}. This stage is to generate reliable initialization of dual-branch network. As mentioned before, target-like images have same annotations inheriting from source images. Hence, we pretrain our dual-branch network jointly, by multi-task learning on the labeled images of both source and target-like domains.
Specifically, the training loss in this stage consists of three terms.
\vspace{-0.05in}
\begin{equation}
\mathcal{L}_{JDP}=\mathcal{L}^{(\mathcal{S+TL})}_{RPN}+\mathcal{L}^{(\mathcal{S})}_{SA}+\mathcal{L}^{(\mathcal{TL})}_{TL}.
\label{eq:lossJDP}
\end{equation}
\vspace{-0.05in}
First, RPN is shared among all the domains to generate domain-invariant feature. We use both source and target-like data to train this module,
i.e.,
$\mathcal{L}^{(\mathcal{S+TL})}_{RPN}=\mathcal{L}_{RPN}(\mathbf{X}^{s},\mathbf{Y}^{s})+\mathcal{L}_{RPN}(\mathbf{X}^{tl},\mathbf{Y}^{tl})$,
where RPN loss contains the RPN classification and regression losses in Faster RCNN\cite{ren2015faster}.
Then, different detection branches are used to learn different domain-specific object knowledge. Hence, we use source and target-like data respectively to train SA and TL branches,
i.e.,
$\mathcal{L}^{(\mathcal{S})}_{SA}=\mathcal{L}_{SA}(\mathbf{X}^{s},\mathbf{Y}^{s})$
and
$\mathcal{L}^{(\mathcal{TL})}_{TL}=\mathcal{L}_{TL}(\mathbf{X}^{tl},\mathbf{Y}^{tl})$
where each branch loss contains the ROI classification and regression losses in Faster RCNN.

\textbf{Cross-Domain Distillation}.
After joint-domain pretraining,
we leverage the well-initialized network to generate pseudo annotations of unlabeled images in the target domain.
In this case,
we can further adjust our network without target-domain ground truth labels.
As shown in Figure \ref{trainingprocess}, this stage is a concise self distillation procedure, where both teacher and student are based on dual branch network.
Specifically, we feed a target-domain image into the fixed and well-trained teacher, which can generate pseudo object annotations from both SA and TL branches. We use NMS to remove the duplicated boxes and then set a threshold to obtain confident box predictions as object annotations of this target image in each branch. Subsequently, we also feed this target image into the learnable student, and train student by pseudo annotations from teacher.
\vspace{-0.05in}
\begin{equation}
\mathcal{L}_{CDD}=\mathcal{L}^{(\mathcal{T})}_{RPN}+\mathcal{L}^{(\mathcal{T})}_{SA}+\mathcal{L}^{(\mathcal{T})}_{TL}.
\label{eq:lossCDD}
\end{equation}
\vspace{-0.05in}
Since pseudo labels $\hat{\mathbf{Y}}^{t}_{SA}$ and $\hat{\mathbf{Y}}^{t}_{TL}$ are from SA and TL branches,
the RPN loss contains two terms
$\mathcal{L}^{(\mathcal{T})}_{RPN}=\mathcal{L}_{RPN}(\mathbf{X}^{t},\hat{\mathbf{Y}}^{t}_{SA})+\mathcal{L}_{RPN}(\mathbf{X}^{t},\hat{\mathbf{Y}}^{t}_{TL})$.
Moreover,
both SA and TL branches are also trained with pseudo labels of target-domain images,
i.e.,
$\mathcal{L}^{(\mathcal{T})}_{SA}=\mathcal{L}_{SA}(\mathbf{X}^{t},\hat{\mathbf{Y}}^{t}_{SA})$
and
$\mathcal{L}^{(\mathcal{T})}_{TL}=\mathcal{L}_{TL}(\mathbf{X}^{t},\hat{\mathbf{Y}}^{t}_{TL})$.
Additionally,
it is important to increase diversity of student to refine teacher afterwards.
As suggested in \cite{liu2021unbiased},
for each target image,
we use its strong augmentation as input of student to predict object boxes,
while using its weak augmentation as input of teacher to provide reliable pseudo annotations.
Finally,
we also use Eq. (\ref{eq:lossJDP}) to train student network with source and target-like images in this stage,
to reduce learning difficulties in two detection branches.

\textbf{Dual-Teacher Refinement}.
To generate more stable pseudo annotations, we refine teacher gradually from student via Exponential Moving Average (EMA)\cite{tarvainen2017mean,liu2021unbiased},
\vspace{-0.05in}
\begin{equation}
\Theta_{teacher}\leftarrow\alpha\Theta_{teacher}+(1-\alpha)\Theta_{student}
\label{eq:lossDTR}
\end{equation}
where
$\Theta_{teacher}$ and $\Theta_{student}$ are the learnable parameters in teacher and student models.
Note that, we perform distillation and refinement in an iterative manner,
which can boost cross domain object detection by mutual learning,
i.e., teacher generates pseudo labels to train student, and student passes what it learns to update teacher. 

Finally, we explain how to train Target Proposal Perceiver in this procedure. We only train it in the last two stages. In the cross-domain distillation stage, we use the pretrained network as teacher, and use the pretrained network with randomly-initialized Target Proposal Perceiver as student. After a number of training iterations in this stage, we can obtain well-trained Target Proposal Perceiver. Subsequently, in the refinement stage, we update teacher from the entire student network where all the modules are fully trained. From then on, distillation and refinement can be iteratively performed without any difficulties. Moreover, TPR is just used in training stage to guide SA branch. With the dual-branch framework, we only use the SA branch teacher to get the detection results during inference. As it has been well refined by student and TL branch.

\section{Experiments}
In this section, we conduct experiments on popular cross domain object detection benchmarks with distinct domain shift, including  Adverse Weather Conditions Adaptation, Synthetic to Real Adaptation and Cross Camera Adaptation.
\subsection{Implementation details}
We adopt Faster R-CNN with the VGG16 and Res50 pretrained on ImageNet\cite{5206848} as the backbone network. The shorter edge of each input image is resized to 600 pixels following the implementation of Faster RCNN  with ROI-alignment\cite{8237584}. The network is trained by SGD\cite{1951A} optimizer with 0.0005 weight decay and 0.9 momentum. The learning rate and maximum training iterations are set as 0.01 and 25000 for all experiments, with 9000 iterations for the joint-domain pretraining stage and 16000 iterations for cross-domain distillation and dual-teacher refinement stage. 
 Follow the\cite{liu2021unbiased}, we use Focal-loss as the classification loss and the strong-weak augmentations are used during our whole training stage. For our proposal cross attention, we set the attention head number L=16. The proposal features are encoded to Key-Query-Value by three FC layers with output dimension=1024. We set the frequency parameter$\beta=0.1$ for the Fourier transform module. The threshold to obtain pseudo annotations of target image is set to 0.7. During the dual-teacher refinement stage, we set the EMA ratio $\alpha=0.9996$ to update teacher model. We use 8 NVIDIA GeForce 1080 Ti GPUs for training in our experiments. Each mini-batch contains 2 images per GPU, one from the source domain and the other from target domain.
\begin{table}[]
\footnotesize
\centering

\caption{The mean Average Precision (mAP) of different models on Foggy Cityscapes validation set for C → F transfer. }
\setlength\tabcolsep{0.9pt}
\begin{tabular}{l|ccccccccccc}
\hline
method & Arch & person & rider & car & truck & bus & train & motor & bike& mAP \\\hline

DA-Faster\cite{8578450} & V16 & 25.0 & 31.0 & 40.5 & 22.1 & 35.3 & 20.2 & 20.0 & 27.1 & 27.6 &  \\
SCDA\cite{8953252} & V16 & 33.5 & 38.0 & 48.5 & 26.5 & 39.0 & 23.3 & 28.0 & 33.6 & 33.8 &  \\
D\&Match\cite{8954025} & V16 & 30.8 & 40.5 & 44.3 & 27.2 & 38.4 & 34.5 & 28.4 & 32.2 & 34.6 &  \\
SWDA\cite{8954336} & V16 & 29.9 & 42.3 & 43.5 & 24.5 & 36.2 & 32.6 & 30.0 & 35.3 & 34.3 &  \\
ICR-CCR\cite{9157382} & V16 & 32.9 & 43.8 & 49.2 & 27.2 & 45.1 & 36.4 & 30.3 & 34.6 & 37.4 &  \\
HTCN\cite{9157147} & V16 & 33.2 & 47.5 & 47.9 & 31.6 & 47.4 & 40.9 & 32.3 & 37.1 & 39.8 &  \\
SAPNet\cite{10.1007/978-3-030-58601-0_29} & V16 & 40.8 & 46.7 & 59.8 & 24.3 & 46.8 & 37.5 & 30.4 & 40.7 & 40.9 \\
ATF\cite{10.1007/978-3-030-58586-0_19}  &V16& 34.6 & 47.0 & 50.0 & 23.7 & 43.3 &38.7 & 33.4 & 38.8 & 38.7\\
CDN\cite{10.1007/978-3-030-58621-8_24}  & V16 & 35.8 & 45.7 & 50.9 & 30.1 & 42.5 & 29.8 & 30.8 & 36.5 & 36.6  \\
UMT\cite{Deng_2021_CVPR} & V16 & 33.0 & 46.7 & 48.6 & 34.1 & 56.5 & 46.8 & 30.4 & 37.3 & 41.7 &  \\
MeGA\cite{VS_2021_CVPR} & V16 & 37.7 & 49.0 & 52.4 & 25.4 & 49.2 & 46.9 & 34.5 & 39.0 & 41.8 &  \\
RPA\cite{Zhang_2021_CVPR} & V16 & 33.4 & 44.3 & 50.1 & 29.9 & 44.8 & 39.1 & 29.9 & 36.3 & 38.5 &  \\\hline 
source only & V16 & 28.5 & 34.2 & 39.9 & 14.7 & 26.3 & 11.4 & 23.4 & 28.3 & 25.8 \\
TDD(ours) &V16 & 39.6 & 47.5 & 55.7 & 33.8 & 47.6 & 42.1 & 37.0 & 41.4 & \textbf{43.1}    \\
ocacle(tgt) & V16 & 39.1 & 44.9 & 56.7 & 33.3 & 50.4 & 34.8 & 32.3 & 39.0 & 41.3   \\
ocacle(src+tgt) & V16 & 39.5 & 47.5 & 58.1 & 34.2 & 49.3 & 41.9 & 36.4 & 41.0 & 43.5 
\\\hline

DA-Faster\cite{8578450} & R50 & 29.2 & 40.4 & 43.4 & 19.7 & 38.3 & 28.5 & 23.7 & 32.7 & 32.0  \\
D\&Match\cite{8954025}  & R50 & 31.8 & 40.5 & 51.0 & 20.9 & 41.8 & 34.3 & 26.6 & 32.4 & 34.9 &  \\
SW-DA\cite{8954336} & R50 & 31.8 & 44.3 & 48.9 & 21.0 & 43.8 & 28.0 & 28.9 & 35.8 & 35.3 &  \\
SC-DA\cite{8953252}  & R50 & 33.8 & 42.1 & 52.1 & 26.8 & 42.5 & 26.5 & 29.2 & 34.5 & 35.9 &  \\
MTOR\cite{8953637} & R50 & 30.6 & 41.4 & 44.0 & 21.9 & 38.6 & 40.6 & 28.3 & 35.6 & 35.1 &  \\
AFAN\cite{9393610}          &R50&42.5&44.6&57.0&26.4&48.0&28.3&33.2&37.1&39.6\\
GPA\cite{9157427} & R50 & 32.9 & 46.7 & 54.1 & 24.7 & 45.7 & 41.1 & 32.4 & 38.7 & 39.5 &  \\
ViSGA\cite{Rezaeianaran_2021_ICCV} & R50 & 38.8 & 45.9 & 57.2 & 29.9 & 50.2 & 51.9 & 31.9 & 40.9 & 43.3 &  \\
SFA\cite{wang2021exploring} & R50 & 46.5 & 48.6 & 62.6 & 25.1 & 46.2 & 29.4 & 28.3 & 44.0   & 41.3 \\
DSS\cite{wang2021domain} & R50 & 50.9 & 57.6 & 61.1 & 35.4 & 50.9 & 36.6 & 38.4 & 51.1 & 47.8  \\
MKT\cite{csaba2021multilevel} & R50 & 43.5 & 52.0   & 63.2 & 34.7 & 52.7 & 45.8 & 37.1 & 49.4 & 47.3 \\
\hline
source only  &R50& 36.9 &36.1 & 44.5 &21.7 &32.3 &9.2 &21.5 &32.4 &28.3 \\
TDD(ours) &R50 &50.7 &53.7 &68.2  &35.1  &53.0 &45.1  &38.9  &49.1 &\textbf{49.2} \\ 
oracle(tgt)   & R50 &50.1  &51.7  &70.1  &33.4  &49.5 &42.8  &37.6 &44.3   &47.4\\
oracle(src+tgt)   &R50 &50.0  &50.2  &69.9  &35.6  &56.3  &47.4  &41.0  &43.4  &49.2 \\\hline 

\end{tabular}
\label{foggy}
\vspace{-0.4cm}
\end{table}

\subsection{Adverse Weather Conditions Adaptation}
\textbf{Datesets.} In this experiment, we use Cityscapes as source domain and Foggy Cityscapes as target domain to implement adaptation under adverse weather conditions (C$\rightarrow$F). Cityscapes\cite{2016The} is a dataset of real urban scenes containing 3,475 images. 2,975 images are used for training and the remaining 500 for validation. Foggy Cityscapes\cite{Sakaridis2018Semantic} is a synthetic dataset generated from the Cityscapes. We use the fog level ($\beta=0.02$) with highest intensity in our experiments. The Cityscapes train set and unlabeled Foggy Cityscapes train set are used for training and the validation set of Foggy Cityscapes is used for evaluation.

\textbf{Results.} The detection results are demonstrated in Table \ref{foggy}. Source only denotes the Faster RCNN model trained with only source domain data. Oracle(tgt) model is trained with labeled target domain data. Oracle(src+tgt) model is trained with labeled data from both source and target domain. Same augmentations are also used for training the oracle models. We compare with the methods implemented with same backbone for fair comparison. For the VGG-based methods, the state-of-the-art MeGA\cite{VS_2021_CVPR} has achieved 41.8\% mAP, while our results show a significant +1.3\% improvement. For the Res50-based methods, we outperform all prior works and get a significant mAP gain of +1.4\%. It is worth noting that our methods show a competitive performance with two oracle models. It proves that our model can perceive target domain knowledge while retaining the useful information of the source domain for discrimination.
\begin{table}[t]
\centering
\footnotesize
\caption{The car Precision (mAP) of different models on Cityscapes validation set for S$\rightarrow$ C and  K $\rightarrow$ C adaptaion.}
\vspace{-0.2cm}
\label{tab:car_on_city}
\setlength\tabcolsep{1pt}
\begin{tabular}{lccc|lccc}
\hline
method &Arch &S$\rightarrow$ C  & K $\rightarrow$ C  &method &Arch  &S$\rightarrow$ C & K $\rightarrow$ C     \\   \hline
DA-Faster\cite{8578450}& V16 & 39.0 &38.5      &DA-Faster\cite{8578450} &R50 & 41.9 &41.8           \\
SCDA\cite{8953252} & V16 & 43.0 &42.5 &SCDA\cite{8953252}   & R50 & 45.1 &43.6\\

SWDA\cite{8954336} & V16 & 47.7 &37.9 &SWDA\cite{8954336}  & R50 & 44.6 &43.2 \\
CoT\cite{10.1007/978-3-030-58523-5_6} &V16 &44.5 &43.6 &GPA\cite{9157427}  & R50 &47.6 &47.9\\
SAPNet\cite{10.1007/978-3-030-58601-0_29} & V16 &44.9  & 43.4  &ViSGA\cite{Rezaeianaran_2021_ICCV}  & R50 &49.3 &47.6 \\
EPM\cite{10.1007/978-3-030-58545-7_42}  & V16 &49.0 & 43.2 &SFA\cite{wang2021exploring} &R50 &52.6 &41.3\\
ATF\cite{10.1007/978-3-030-58586-0_19}  & V16 &42.8 & 42.1 &D\&Match\cite{8954025}  & R50 &43.9 &42.7\\ 
MeGA\cite{VS_2021_CVPR} & V16 & 44.8 &43.0 &DSS\cite{wang2021domain} & R50 &44.5 &42.7 \\

RPA\cite{Zhang_2021_CVPR} & V16 & 45.7 &- &MKT\cite{csaba2021multilevel} & R50 &50.2 &44.3\\
C2F\cite{9156464} &V16 &43.8 &- &AFAN\cite{9393610} & R50 &45.5 &- 
 \\
UMT\cite{Deng_2021_CVPR} & V16 & 43.1 &- &MTOR\cite{8953637}  & R50 & 46.6 &- \\

\hline
source only	&V16 &37.8 &30.2 &source only&R50 &42.8 &32.5 \\
TDD(ours) & V16 & \textbf{53.4} &\textbf{47.4} &TDD(ours) & R50 &\textbf{63.3} & \textbf{49.8} \\
oracle(tgt) & V16 & 60.0 & 60.0 &oracle(tgt) &R50 &75.9 &75.9\\
oracle(src+tgt)&V16 &60.1 &62.5 &oracle(src+tgt) &R50 &76.4 &75.8 \\\hline
\end{tabular}
\vspace{-0.2cm}
\end{table}

\begin{table}[]
\centering
\footnotesize
\setlength\tabcolsep{1pt}
\caption{The mean Average Precision (mAP) of different models on BDD100k daytime validation set for C → B transfer. }
\vspace{-0.2cm}
\begin{tabular}{l|ccccccccccc}
\hline
method &Arch & person & rider & car & truck & bus  & motor & bicycle & mAP \\\hline
DA-Faster\cite{8578450}  & V16 &26.9  &22.1  &44.7  &17.4  &16.7  &17.1  &18.8  &23.4  \\

SWDA\cite{8954336} &V16  &30.2  &29.5  &45.7  &15.2  &18.4   &17.1 &21.2 &25.3  \\
ICR-CCR\cite{9157382}  &V16 &31.4  &31.3  &46.3  &19.5  &18.9   &17.3 &23.8 &26.9  \\  \hline
source only  &V16 & 29.3 & 28.2 & 45.7 & 15.5 & 16.6 & 16.0 & 22.1 & 24.8\\
TDD(ours) &V16 & 39.6 & 38.9 & 53.9 & 24.1 & 25.5 & 24.5 & 28.8  &\textbf{33.6}\\ 
oracle(tgt) & V16 & 39.7 & 35.9 & 57.9 & 47.1 & 48.0 & 32.3 & 33.0 & 42.0 \\
oracle(src+tgt) & V16 & 39.6 & 39.2 & 59.4 & 45.6 & 48.0 & 31.0 & 33.8 & 42.4\\
\hline
source only  & R50 &50.4  &33.3  &67.4  &18.1  &20.8   &19.6  &28.9  &34.1  \\

TDD(ours) &R50  &57.9  &47.4  &74.5  &31.5 &27.5 &  32.0 &36.5 &\textbf{43.9}  \\ 

oracle(tgt)& R50 &68.0  &52.0  &83.7  &61.2  &61.6   &44.9  &49.9   &60.2  \\
oracle(src+tgt)  & R50 &69.5 & 54.1 & 84.4 & 61.1 & 61.5 & 43.8 & 53.2 & 61.1\\ \hline
\end{tabular}
\label{tab:bdd_result}
\vspace{-0.5cm}
\end{table}

\vspace{-0.5cm}
\subsection{Synthetic to Real Adaptation}

\textbf{Datesets.} In this experiment, the model is adapted from synthetic data to real world examples. Sim10k is used as source domain dataset and Cityscapes represents target domain (S$\rightarrow$ C). SIM10K\cite{2017Driving} is a simulated dataset containing 10,000 images. We train the detector only on the common class “car”. The whole dataset Sim10k and unlabeled train set of Cityscapes is used for training and the validation set of Cityscapes is used for evaluation.

\textbf{Results.} The results of car AP are reported in Table \ref{tab:car_on_city}. We can see our proposed TDD methods can achieve the state-of-the-art performance between two dissimilar domains. It outperforms VGG-based EPM\cite{10.1007/978-3-030-58545-7_42} by +4.4\% and Res50-based SFA\cite{wang2021exploring} +10.7\%, which 
shows a stable ability of our methods to tackle domain adaptation problems.

\subsection{Cross Camera Adaption}
\textbf{Datesets.} We conduct on two cross camera adapataion experiments involving KITTI\cite{2012Are}, Cityscapes and BDD100k\cite{9156329} datasets. In the first experiment, we adapt from KITTI to Cityscapes, where only the category car is used for evaluation (K$\rightarrow$C). KITTI is a similar scene dataset to Cityscapes except that KITTI has different camera setup. It consists of 7,481 labeled images for training. In the second experiment, we adapt from Cityscapes to BDD100K (C$\rightarrow$B), which is a more challenging setting with more categories and scenes. The daytime subset of BDD100k are used as our target domain, including 36,278 training and 5,258 validation images.

\textbf{Results.} The KITTI adaptation results are shown in Table \ref{tab:car_on_city}. We outperform the sota VGG-based approach by 3.8\% and R50-based approach by 1.9\%.
Meanwhile, the results on BDD100K are summarized in Table \ref{tab:bdd_result}. Our method surpasses all the previous works with a large margin. This demonstrates that our method performs well under more complex situation. We also observe an obvious improvement with R50 backbone, increasing the source only results by 9.8\%. It further verifies the robustness of our methods.
\begin{table}[]
\centering
\footnotesize
\caption{Dual Branch Structure}
\vspace{-0.2cm}
\begin{tabular}{l|lll|lll}
\hline
Structure        & S & T & TL & C$\rightarrow$ F & S$\rightarrow$ C & C$\rightarrow$ B \\\hline
       &  \checkmark &   &    & 34.8    & 48.3  & 34.3    \\
Single &  \checkmark &  \checkmark &    & 41.2    & 59.0     & 38.9      \\
       &  \checkmark &  \checkmark &  \checkmark  & 47.4    & 61.1   & 39.4    \\\hline
Dual   &  \checkmark &  \checkmark &  \checkmark  & \textbf{48.3}    &\textbf{62.6}  & \textbf{42.2}   \\\hline
\end{tabular}
\label{tab:ablation_of_dual_branch}
\vspace{-0.2cm}
\end{table}

\begin{table}[]
\centering
\footnotesize
\caption{Multi Head Proposal Cross Attention}
\vspace{-0.2cm}

\begin{tabular}{l|ccc}
\hline
Target Proposal Perceiver  & C$\rightarrow$ F & S$\rightarrow$ C & C$\rightarrow$ B \\\hline
without              & 48.3 & 62.6 & 42.2 \\
with                 & \textbf{49.2} &\textbf{63.3}& \textbf{43.9}  \\ \hline
Self-Attention       & 46.8 & 61.0 & 40.6 \\ 
Sym Cross-Attention  & 48.1 & 62.4 & 43.7 \\
Asym Cross-Attention & \textbf{49.2} &\textbf{63.3}& \textbf{43.9} \\ \hline
\end{tabular}
\label{ablation on MHPCA}
\vspace{-0.2cm}
\end{table}

\begin{table}[]
\centering
\footnotesize
\caption{Dual-Branch Self Distillation Procedure}
\vspace{-0.2cm}
\begin{tabular}{l|ccc}
\hline
Dual-Branch Self Distillation & C$\rightarrow$ F & S$\rightarrow$ C & C$\rightarrow$ B \\\hline 
JDP  &37.4 &56.7 &37.5 \\
JDP+CDD & 44.1    & 62.1   & 42.7 \\
JDP+CDD+DTR & \textbf{49.2}    & \textbf{63.3}   & \textbf{43.9}\\ \hline
Refine $\alpha$=0.96   &  39.3    & 59.1   & 28.7 \\
Refine $\alpha$=0.996  & 48.4    & \textbf{63.6}   & 41.5   \\
Refine $\alpha$=0.9996 & \textbf{49.2}    & 63.3   & \textbf{43.9}   
\\\hline
\end{tabular}
\label{process}
\vspace{-0.2cm}
\end{table}

\subsection{Ablation Studies and Analysis}
To verify designs in our network, we conduct a set of ablation studies on the Res50 backbone.
\begin{figure*}[t]
  \centering
  \includegraphics[width=0.9\linewidth]{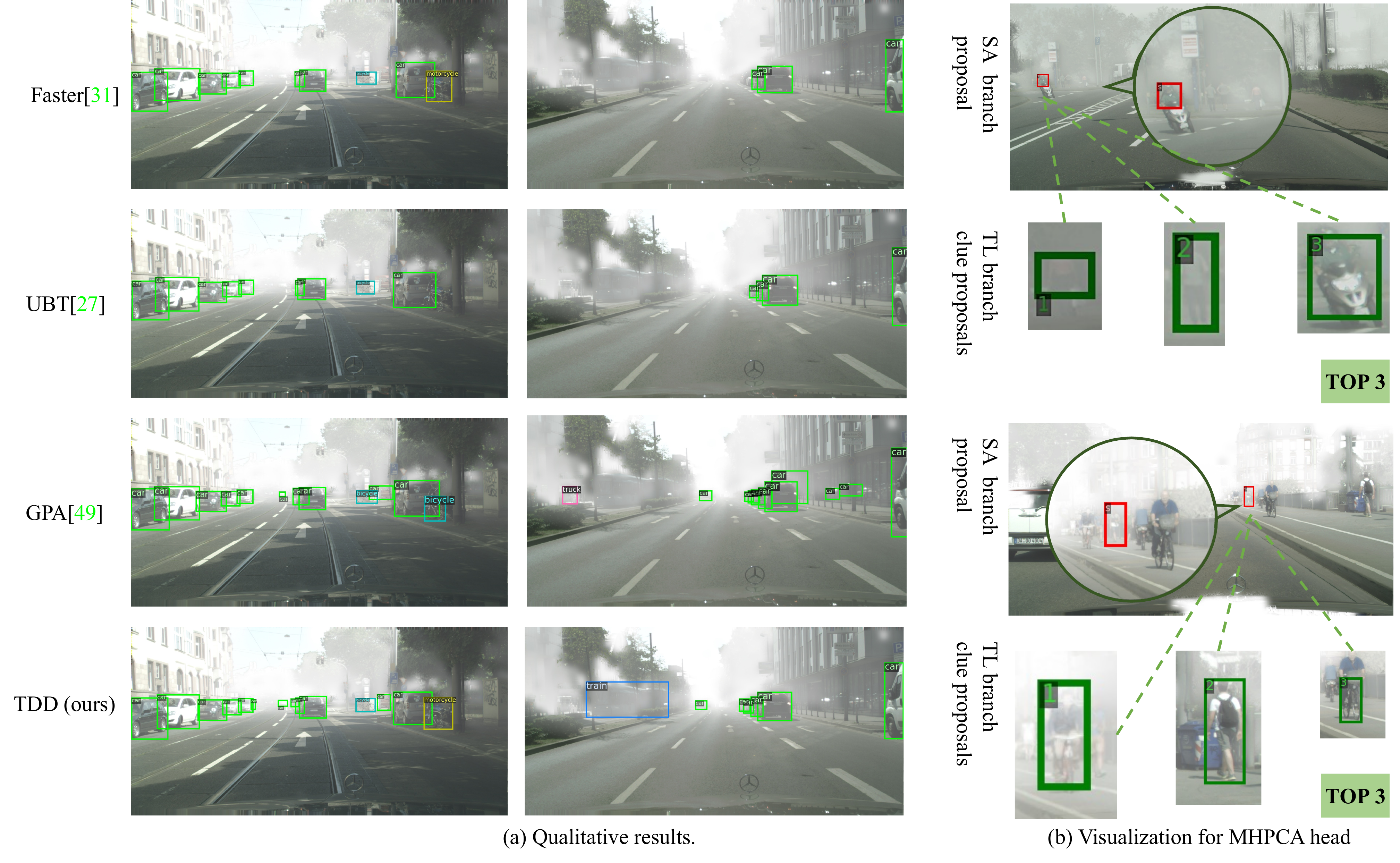}
  \vspace{-0.3cm}
   \caption{(a): Qualitative detection results of on the $C\rightarrow F$ scenario for different models. We set confidence thresh=0.6 for visualization. (b): Top 3 most relevant clue proposals in TL branch found by our MHPCA.}
  \vspace{-0.5cm}
\label{fig:vis}
\end{figure*}

\textbf{Dual branch}. 
To validate the effectiveness of our dual branch structure, we conduct a set of ablation studies with images from different domains. Table \ref{tab:ablation_of_dual_branch} shows the results of different experiments. When implemented with a single branch, target domain images are supervised by the pseudo annotations generated by the single teacher branch, while the target-like images were feed to the network paired with source images. The proposed Target-proposal-perceiver is not used in this dual-branch structure to fairly compare with the single-branch experiments. We can observed that the model performance improved step by step with the target and target-like images participating in the training. This verifies our motivation that data from each domain is useful. The dual-branch experiment outperforms all the single branch methods, which demonstrates that our dual-branch distillation framework can effectively retain the useful source domain knowledge and explore target domain information simultaneously.

\textbf{Multi Head Proposal Cross Attention.} We implement the MHPCA to guide the source adaptive branch to learn knowledge closer to the target domain with the help of target-like domain branch. Table \ref{ablation on MHPCA} shows the effectiveness of our MHPCA module. First, we can see a significant improvement with the MHPCA module added. Moreover, to validate that domain difference between two branches matters for our attention model, we also experiment with self-attention.  Besides, to explore the guide manner between two branch, we also add a cross attention head in the target-like branch. Sym Cross-Attention means that a same cross-attention module is added on both the source-adaptive and target-like branch. While Asym Cross-Attention refers to our TDD methods which equip the SA branch with the cross attention module. Our asymmetric TDD performs best in these three manners. It also confirms that in our framework, the cross attention manner is needed for the SA branch due to the lack of target domain knowledge.

\textbf{Dual-Branch Self Distillation.} We do ablation studies to verify the effectiveness of our dual-branch self distillation procedure, which is composed of Joint-Domain Pretraining (JDP), Cross-Domain Distillation (CDD), and Dual Teacher Refinement (DTR) steps. We see from Table \ref{process} that all of the three steps in our method improve former step results. We also experiment with different EMA rate $\alpha$ in dual-teacher refinement stage. The smaller the value of $\alpha$, the more information teacher receives from the target image during the refine stage. When $\alpha$ is set to be small (e.g., 0.96), the model performance drops significantly. Additionally, when $\alpha=1$, the teacher is not refined which is JDP+CDD in Table \ref{process}. All these show the teacher model should be updated gradually. A reasonable rate is needed to impart the target domain knowledge learned by student to teacher.

\textbf{Qualitative results.} We show the detection results of Faster\cite{ren2015faster}, GPA\cite{9157427}, UBT\cite{liu2021unbiased} and our TDD in Figure \ref{fig:vis} (a). We can see that many objects can not be detected by the Faster RCNN and UBT due to the heavy fog, while the GPA attempts to capture objects in the fog but gives wrong prediction. Our TDD can localize and classify objects more accurately. We also visualize the working mechanism for our cross-domain MHPCA module. For a SA branch proposal, our attention head can discover useful contextual proposal features in TL branch as clues for detection. As the top image shown in Figure \ref{fig:vis}(b), a rider is classified with the guidance of a motorcycle and two person proposals. 

\section{Conclusion}
In this work, we propose a novel Target-perceived Dual branch Distillation framework. Through a target proposal perceiver and our dual-branch self distillation procedure, we tackle domain shift and label deficiency together in cross domain object detection. Extensive experiments are conducted on multiple benchmarks, and the results clearly show that our TDD surpasses the existing state-of-the-art models.
\textbf{Acknowledgement:} This work is partially supported by the National Natural Science Foundation of China (61876176,U1813218), the Joint Lab of CASHK, Guangdong NSF Project(No. 2020B1515120085),the Shenzhen Research Program(RCJC20200714114557087),  the Shanghai Committee of Science and Technology, China (Grant No. 21DZ1100100).

{\small
\bibliographystyle{ieee_fullname}
\bibliography{egbib}

\begin{thebibliography}{10}\itemsep=-1pt

\bibitem{8953637}
Qi Cai, Yingwei Pan, Chong-Wah Ngo, Xinmei Tian, Lingyu Duan, and Ting Yao.
\newblock Exploring object relation in mean teacher for cross-domain detection.
\newblock In {\em 2019 IEEE/CVF Conference on Computer Vision and Pattern
  Recognition (CVPR)}, pages 11449--11458, 2019.

\bibitem{cai2018cascade}
Zhaowei Cai and Nuno Vasconcelos.
\newblock Cascade r-cnn: Delving into high quality object detection.
\newblock In {\em Proceedings of the IEEE conference on computer vision and
  pattern recognition}, pages 6154--6162, 2018.

\bibitem{carion2020end}
Nicolas Carion, Francisco Massa, Gabriel Synnaeve, Nicolas Usunier, Alexander
  Kirillov, and Sergey Zagoruyko.
\newblock End-to-end object detection with transformers.
\newblock In {\em European Conference on Computer Vision}, pages 213--229.
  Springer, 2020.

\bibitem{9157147}
Chaoqi Chen, Zebiao Zheng, Xinghao Ding, Yue Huang, and Qi Dou.
\newblock Harmonizing transferability and discriminability for adapting object
  detectors.
\newblock In {\em 2020 IEEE/CVF Conference on Computer Vision and Pattern
  Recognition (CVPR)}, pages 8866--8875, 2020.

\bibitem{8578450}
Yuhua Chen, Wen Li, Christos Sakaridis, Dengxin Dai, and Luc Van~Gool.
\newblock Domain adaptive faster r-cnn for object detection in the wild.
\newblock In {\em 2018 IEEE/CVF Conference on Computer Vision and Pattern
  Recognition}, pages 3339--3348, 2018.

\bibitem{2016The}
M. Cordts, M. Omran, S. Ramos, T. Rehfeld, and B. Schiele.
\newblock The cityscapes dataset for semantic urban scene understanding.
\newblock In {\em 2016 IEEE Conference on Computer Vision and Pattern
  Recognition (CVPR)}, 2016.

\bibitem{csaba2021multilevel}
Botos Csaba, Xiaojuan Qi, Arslan Chaudhry, Puneet Dokania, and Philip Torr.
\newblock Multilevel knowledge transfer for cross-domain object detection.
\newblock {\em arXiv preprint arXiv:2108.00977}, 2021.

\bibitem{5206848}
Jia Deng, Wei Dong, Richard Socher, Li-Jia Li, Kai Li, and Li Fei-Fei.
\newblock Imagenet: A large-scale hierarchical image database.
\newblock In {\em 2009 IEEE Conference on Computer Vision and Pattern
  Recognition}, pages 248--255, 2009.

\bibitem{Deng_2021_CVPR}
Jinhong Deng, Wen Li, Yuhua Chen, and Lixin Duan.
\newblock Unbiased mean teacher for cross-domain object detection.
\newblock In {\em Proceedings of the IEEE/CVF Conference on Computer Vision and
  Pattern Recognition (CVPR)}, pages 4091--4101, June 2021.

\bibitem{duan2019centernet}
Kaiwen Duan, Song Bai, Lingxi Xie, Honggang Qi, Qingming Huang, and Qi Tian.
\newblock Centernet: Keypoint triplets for object detection.
\newblock In {\em Proceedings of the IEEE/CVF International Conference on
  Computer Vision}, pages 6569--6578, 2019.

\bibitem{2012Are}
A. Geiger, P. Lenz, and R. Urtasun.
\newblock Are we ready for autonomous driving? the kitti vision benchmark
  suite.
\newblock In {\em IEEE Conference on Computer Vision \& Pattern Recognition},
  2012.

\bibitem{7410526}
Ross Girshick.
\newblock Fast r-cnn.
\newblock In {\em 2015 IEEE International Conference on Computer Vision
  (ICCV)}, pages 1440--1448, 2015.

\bibitem{8237584}
Kaiming He, Georgia Gkioxari, Piotr Dollár, and Ross Girshick.
\newblock Mask r-cnn.
\newblock In {\em 2017 IEEE International Conference on Computer Vision
  (ICCV)}, pages 2980--2988, 2017.

\bibitem{he2015spatial}
Kaiming He, Xiangyu Zhang, Shaoqing Ren, and Jian Sun.
\newblock Spatial pyramid pooling in deep convolutional networks for visual
  recognition.
\newblock {\em IEEE transactions on pattern analysis and machine intelligence},
  37(9):1904--1916, 2015.

\bibitem{9010003}
Zhenwei He and Lei Zhang.
\newblock Multi-adversarial faster-rcnn for unrestricted object detection.
\newblock In {\em 2019 IEEE/CVF International Conference on Computer Vision
  (ICCV)}, pages 6667--6676, 2019.

\bibitem{10.1007/978-3-030-58586-0_19}
Zhenwei He and Lei Zhang.
\newblock Domain adaptive object detection via asymmetric tri-way faster-rcnn.
\newblock In Andrea Vedaldi, Horst Bischof, Thomas Brox, and Jan-Michael Frahm,
  editors, {\em Computer Vision -- ECCV 2020}, pages 309--324, Cham, 2020.
  Springer International Publishing.

\bibitem{10.1007/978-3-030-58545-7_42}
Cheng-Chun Hsu, Yi-Hsuan Tsai, Yen-Yu Lin, and Ming-Hsuan Yang.
\newblock Every pixel matters: Center-aware feature alignment for domain
  adaptive object detector.
\newblock In Andrea Vedaldi, Horst Bischof, Thomas Brox, and Jan-Michael Frahm,
  editors, {\em Computer Vision -- ECCV 2020}, pages 733--748, Cham, 2020.
  Springer International Publishing.

\bibitem{hu2018relation}
Han Hu, Jiayuan Gu, Zheng Zhang, Jifeng Dai, and Yichen Wei.
\newblock Relation networks for object detection.
\newblock In {\em Proceedings of the IEEE conference on computer vision and
  pattern recognition}, pages 3588--3597, 2018.

\bibitem{DBLP:journals/corr/abs-2103-03206}
Andrew Jaegle, Felix Gimeno, Andrew Brock, Andrew Zisserman, Oriol Vinyals, and
  Jo{\~{a}}o Carreira.
\newblock Perceiver: General perception with iterative attention.
\newblock {\em CoRR}, abs/2103.03206, 2021.

\bibitem{jeong2019consistency}
Jisoo Jeong, Seungeui Lee, Jeesoo Kim, and Nojun Kwak.
\newblock Consistency-based semi-supervised learning for object detection.
\newblock {\em Advances in neural information processing systems},
  32:10759--10768, 2019.

\bibitem{2017Driving}
M. Johnson-Roberson, C. Barto, R. Mehta, S.~N. Sridhar, and R. Vasudevan.
\newblock Driving in the matrix: Can virtual worlds replace human-generated
  annotations for real world tasks?
\newblock In {\em IEEE International Conference on Robotics \& Automation},
  2017.

\bibitem{9008383}
Mehran Khodabandeh, Arash Vahdat, Mani Ranjbar, and William Macready.
\newblock A robust learning approach to domain adaptive object detection.
\newblock In {\em 2019 IEEE/CVF International Conference on Computer Vision
  (ICCV)}, pages 480--490, 2019.

\bibitem{9010241}
Seunghyeon Kim, Jaehoon Choi, Taekyung Kim, and Changick Kim.
\newblock Self-training and adversarial background regularization for
  unsupervised domain adaptive one-stage object detection.
\newblock In {\em 2019 IEEE/CVF International Conference on Computer Vision
  (ICCV)}, pages 6091--6100, 2019.

\bibitem{8954025}
Taekyung Kim, Minki Jeong, Seunghyeon Kim, Seokeon Choi, and Changick Kim.
\newblock Diversify and match: A domain adaptive representation learning
  paradigm for object detection.
\newblock In {\em 2019 IEEE/CVF Conference on Computer Vision and Pattern
  Recognition (CVPR)}, pages 12448--12457, 2019.

\bibitem{10.1007/978-3-030-58601-0_29}
Congcong Li, Dawei Du, Libo Zhang, Longyin Wen, Tiejian Luo, Yanjun Wu, and
  Pengfei Zhu.
\newblock Spatial attention pyramid network for unsupervised domain adaptation.
\newblock In Andrea Vedaldi, Horst Bischof, Thomas Brox, and Jan-Michael Frahm,
  editors, {\em Computer Vision -- ECCV 2020}, pages 481--497, Cham, 2020.
  Springer International Publishing.

\bibitem{lin2017focal}
Tsung-Yi Lin, Priya Goyal, Ross Girshick, Kaiming He, and Piotr Doll{\'a}r.
\newblock Focal loss for dense object detection.
\newblock In {\em Proceedings of the IEEE international conference on computer
  vision}, pages 2980--2988, 2017.

\bibitem{liu2021unbiased}
Yen-Cheng Liu, Chih-Yao Ma, Zijian He, Chia-Wen Kuo, Kan Chen, Peizhao Zhang,
  Bichen Wu, Zsolt Kira, and Peter Vajda.
\newblock Unbiased teacher for semi-supervised object detection.
\newblock In {\em International Conference on Learning Representations}, 2021.

\bibitem{7780460}
Joseph Redmon, Santosh Divvala, Ross Girshick, and Ali Farhadi.
\newblock You only look once: Unified, real-time object detection.
\newblock In {\em 2016 IEEE Conference on Computer Vision and Pattern
  Recognition (CVPR)}, pages 779--788, 2016.

\bibitem{redmon2017yolo9000}
Joseph Redmon and Ali Farhadi.
\newblock Yolo9000: better, faster, stronger.
\newblock In {\em Proceedings of the IEEE conference on computer vision and
  pattern recognition}, pages 7263--7271, 2017.

\bibitem{redmon2018yolov3}
Joseph Redmon and Ali Farhadi.
\newblock Yolov3: An incremental improvement.
\newblock {\em arXiv preprint arXiv:1804.02767}, 2018.

\bibitem{ren2015faster}
Shaoqing Ren, Kaiming He, Ross Girshick, and Jian Sun.
\newblock Faster r-cnn: Towards real-time object detection with region proposal
  networks.
\newblock {\em Advances in neural information processing systems}, 28:91--99,
  2015.

\bibitem{Rezaeianaran_2021_ICCV}
Farzaneh Rezaeianaran, Rakshith Shetty, Rahaf Aljundi, Daniel~Olmeda Reino,
  Shanshan Zhang, and Bernt Schiele.
\newblock Seeking similarities over differences: Similarity-based domain
  alignment for adaptive object detection.
\newblock In {\em Proceedings of the IEEE/CVF International Conference on
  Computer Vision (ICCV)}, pages 9204--9213, October 2021.

\bibitem{1951A}
Herbert Robbins and Sutton Monro.
\newblock A stochastic approximation method.
\newblock {\em Annals of Mathematical Statistics}, 22(3):400--407, 1951.

\bibitem{8954336}
Kuniaki Saito, Yoshitaka Ushiku, Tatsuya Harada, and Kate Saenko.
\newblock Strong-weak distribution alignment for adaptive object detection.
\newblock In {\em 2019 IEEE/CVF Conference on Computer Vision and Pattern
  Recognition (CVPR)}, pages 6949--6958, 2019.

\bibitem{Sakaridis2018Semantic}
Sakaridis, Christos, Dai, Dengxin, Van, Gool, and Luc.
\newblock Semantic foggy scene understanding with synthetic data.
\newblock {\em International Journal of Computer Vision}, 2018.

\bibitem{shen2017dsod}
Zhiqiang Shen, Zhuang Liu, Jianguo Li, Yu-Gang Jiang, Yurong Chen, and
  Xiangyang Xue.
\newblock Dsod: Learning deeply supervised object detectors from scratch.
\newblock In {\em Proceedings of the IEEE international conference on computer
  vision}, pages 1919--1927, 2017.

\bibitem{sohn2020simple}
Kihyuk Sohn, Zizhao Zhang, Chun-Liang Li, Han Zhang, Chen-Yu Lee, and Tomas
  Pfister.
\newblock A simple semi-supervised learning framework for object detection.
\newblock {\em arXiv preprint arXiv:2005.04757}, 2020.

\bibitem{10.1007/978-3-030-58621-8_24}
Peng Su, Kun Wang, Xingyu Zeng, Shixiang Tang, Dapeng Chen, Di Qiu, and
  Xiaogang Wang.
\newblock Adapting object detectors with conditional domain normalization.
\newblock In Andrea Vedaldi, Horst Bischof, Thomas Brox, and Jan-Michael Frahm,
  editors, {\em Computer Vision -- ECCV 2020}, pages 403--419, Cham, 2020.
  Springer International Publishing.

\bibitem{tarvainen2017mean}
Antti Tarvainen and Harri Valpola.
\newblock Mean teachers are better role models: Weight-averaged consistency
  targets improve semi-supervised deep learning results.
\newblock {\em arXiv preprint arXiv:1703.01780}, 2017.

\bibitem{tian2019fcos}
Zhi Tian, Chunhua Shen, Hao Chen, and Tong He.
\newblock Fcos: Fully convolutional one-stage object detection.
\newblock In {\em Proceedings of the IEEE/CVF international conference on
  computer vision}, pages 9627--9636, 2019.

\bibitem{vaswani2017attention}
Ashish Vaswani, Noam Shazeer, Niki Parmar, Jakob Uszkoreit, Llion Jones,
  Aidan~N Gomez, {\L}ukasz Kaiser, and Illia Polosukhin.
\newblock Attention is all you need.
\newblock In {\em Advances in neural information processing systems}, pages
  5998--6008, 2017.

\bibitem{VS_2021_CVPR}
Vibashan VS, Vikram Gupta, Poojan Oza, Vishwanath~A. Sindagi, and Vishal~M.
  Patel.
\newblock Mega-cda: Memory guided attention for category-aware unsupervised
  domain adaptive object detection.
\newblock In {\em Proceedings of the IEEE/CVF Conference on Computer Vision and
  Pattern Recognition (CVPR)}, pages 4516--4526, June 2021.

\bibitem{9393610}
Hongsong Wang, Shengcai Liao, and Ling Shao.
\newblock Afan: Augmented feature alignment network for cross-domain object
  detection.
\newblock {\em IEEE Transactions on Image Processing}, 30:4046--4056, 2021.

\bibitem{wang2021exploring}
Wen Wang, Yang Cao, Jing Zhang, Fengxiang He, Zheng-Jun Zha, Yonggang Wen, and
  Dacheng Tao.
\newblock Exploring sequence feature alignment for domain adaptive detection
  transformers.
\newblock In {\em Proceedings of the 29th ACM International Conference on
  Multimedia}, pages 1730--1738, 2021.

\bibitem{wang2021pyramid}
Wenhai Wang, Enze Xie, Xiang Li, Deng-Ping Fan, Kaitao Song, Ding Liang, Tong
  Lu, Ping Luo, and Ling Shao.
\newblock Pyramid vision transformer: A versatile backbone for dense prediction
  without convolutions.
\newblock {\em arXiv preprint arXiv:2102.12122}, 2021.

\bibitem{wang2021domain}
Yu Wang, Rui Zhang, Shuo Zhang, Miao Li, YangYang Xia, XiShan Zhang, and ShaoLi
  Liu.
\newblock Domain-specific suppression for adaptive object detection.
\newblock In {\em Proceedings of the IEEE/CVF Conference on Computer Vision and
  Pattern Recognition}, pages 9603--9612, 2021.

\bibitem{9022084}
Rongchang Xie, Fei Yu, Jiachao Wang, Yizhou Wang, and Li Zhang.
\newblock Multi-level domain adaptive learning for cross-domain detection.
\newblock In {\em 2019 IEEE/CVF International Conference on Computer Vision
  Workshop (ICCVW)}, pages 3213--3219, 2019.

\bibitem{9157382}
Chang-Dong Xu, Xing-Ran Zhao, Xin Jin, and Xiu-Shen Wei.
\newblock Exploring categorical regularization for domain adaptive object
  detection.
\newblock In {\em 2020 IEEE/CVF Conference on Computer Vision and Pattern
  Recognition (CVPR)}, pages 11721--11730, 2020.

\bibitem{9157427}
Minghao Xu, Hang Wang, Bingbing Ni, Qi Tian, and Wenjun Zhang.
\newblock Cross-domain detection via graph-induced prototype alignment.
\newblock In {\em 2020 IEEE/CVF Conference on Computer Vision and Pattern
  Recognition (CVPR)}, pages 12352--12361, 2020.

\bibitem{9157228}
Yanchao Yang and Stefano Soatto.
\newblock Fda: Fourier domain adaptation for semantic segmentation.
\newblock In {\em 2020 IEEE/CVF Conference on Computer Vision and Pattern
  Recognition (CVPR)}, pages 4084--4094, 2020.

\bibitem{yang2019reppoints}
Ze Yang, Shaohui Liu, Han Hu, Liwei Wang, and Stephen Lin.
\newblock Reppoints: Point set representation for object detection.
\newblock In {\em Proceedings of the IEEE/CVF International Conference on
  Computer Vision}, pages 9657--9666, 2019.

\bibitem{9156329}
Fisher Yu, Haofeng Chen, Xin Wang, Wenqi Xian, Yingying Chen, Fangchen Liu,
  Vashisht Madhavan, and Trevor Darrell.
\newblock Bdd100k: A diverse driving dataset for heterogeneous multitask
  learning.
\newblock In {\em 2020 IEEE/CVF Conference on Computer Vision and Pattern
  Recognition (CVPR)}, pages 2633--2642, 2020.

\bibitem{zhang2020bridging}
Shifeng Zhang, Cheng Chi, Yongqiang Yao, Zhen Lei, and Stan~Z Li.
\newblock Bridging the gap between anchor-based and anchor-free detection via
  adaptive training sample selection.
\newblock In {\em Proceedings of the IEEE/CVF conference on computer vision and
  pattern recognition}, pages 9759--9768, 2020.

\bibitem{Zhang_2021_CVPR}
Yixin Zhang, Zilei Wang, and Yushi Mao.
\newblock Rpn prototype alignment for domain adaptive object detector.
\newblock In {\em Proceedings of the IEEE/CVF Conference on Computer Vision and
  Pattern Recognition (CVPR)}, pages 12425--12434, June 2021.

\bibitem{10.1007/978-3-030-58523-5_6}
Ganlong Zhao, Guanbin Li, Ruijia Xu, and Liang Lin.
\newblock Collaborative training between region proposal localization and
  classification for domain adaptive object detection.
\newblock In Andrea Vedaldi, Horst Bischof, Thomas Brox, and Jan-Michael Frahm,
  editors, {\em Computer Vision -- ECCV 2020}, pages 86--102, Cham, 2020.
  Springer International Publishing.

\bibitem{9156464}
Yangtao Zheng, Di Huang, Songtao Liu, and Yunhong Wang.
\newblock Cross-domain object detection through coarse-to-fine feature
  adaptation.
\newblock In {\em 2020 IEEE/CVF Conference on Computer Vision and Pattern
  Recognition (CVPR)}, pages 13763--13772, 2020.

\bibitem{8953252}
Xinge Zhu, Jiangmiao Pang, Ceyuan Yang, Jianping Shi, and Dahua Lin.
\newblock Adapting object detectors via selective cross-domain alignment.
\newblock In {\em 2019 IEEE/CVF Conference on Computer Vision and Pattern
  Recognition (CVPR)}, pages 687--696, 2019.

\bibitem{zhu2020deformable}
Xizhou Zhu, Weijie Su, Lewei Lu, Bin Li, Xiaogang Wang, and Jifeng Dai.
\newblock Deformable detr: Deformable transformers for end-to-end object
  detection.
\newblock {\em arXiv preprint arXiv:2010.04159}, 2020.

\end{thebibliography}
}

\end{document}